\documentclass{article}%
\usepackage[T1]{fontenc}%
\usepackage[utf8]{inputenc}%
\usepackage{lmodern}%
\usepackage{textcomp}%
\usepackage{lastpage}%
\usepackage{caption}%
\usepackage[section]{placeins}  
\usepackage{geometry}%
\usepackage{multicol}%
\usepackage{graphicx}%
\usepackage{lipsum}%
\usepackage{booktabs}%
\usepackage{ragged2e}%
\title{Unsupervised Machine learning methods for city vitality index }%
\author{Jean-Sébastien Dessureault, Jonathan Simard, and Daniel Massicotte}%
\date{}
\begin{document}%
\normalsize%
\maketitle%

\noindent Universit\'{e} du Qu\'{e}bec \`{a} Trois-Rivi\`{e}res, Department of Electrical and Computer Engineering, \newline
\noindent 3351, Boul. des Forges, Trois-Rivi\`{e}res, Qu\'{e}bec, Canada\newline
\noindent Laboratoire des Signaux et Syst\`{e}mes Int\'{e}gr\'{e}s\newline
\noindent $\mathrm{\{}$sebastien.dessureault, jonathan.simard, daniel.massicotte$\mathrm{\}}$@uqtr.ca\newline\newline


\justify%
\section*{ABSTRACT}%
\label{sec:ABSTRACT}%
This paper concerns the challenge to evaluate and predict a district vitality index (VI) over the years. There is no standard method to do it, and it is even more complicated to do it retroactively in the last decades. Although, it is essential to evaluate and learn features of the past to predict a VI in the future. This paper proposes a method to evaluate such a VI, based on a \textit{k}-mean clustering algorithm. The meta parameters of this unsupervised machine learning technique are optimized by a genetic algorithm method. Based on the resulting clusters and VI, a linear regression is applied to predict the VI of each district of a city. The weights of each feature used in the clustering are calculated using a random forest regressor algorithm. This method can be a powerful insight for urbanists and inspire the redaction of a city plan in the smart city context.

\begin{multicols}{2}%
\section{Introduction}%
\label{sec:Introduction}%
\noindent Cities are constantly evaluating. Too often, several districts in a city has been forgotten for years and without warning, they are devitalized. It is often too late to act. People and businesses are leaving this district because of many factors such as the disuse of the houses and the buildings, the bad economic activities, the criminality rate, and so on. Even though it is easy to note when a district is already devitalized, urbanists do not have some good tools to predict which district will be devitalized, and when. 

\noindent Historically, the firsts relevant publications were made between 1960 and 1980. The works of Donald Appleyard for instance with a study named "Styles and methods of structuring a city" \cite{appleyard_styles_1970} was aiming to explain different city patterns according to some features like the level of education, the age, the sex and so on.  Having back then very few processing capacities, those works was preparing a new numeric era for urbanism. 

\noindent In the late 70s, a paper \cite{kuipers_modeling_1978} introduced some concept that will become a Geographical Information System (GIS). There were many publications at the beginning of the 80's studying those new GIS. Muehrcke in \cite{Muehrcke_1990} made a good review of the literature at this time. Since then, the GIS are a very useful tool to every system that must manage geographical data. 

\noindent The concept of smart city came later at the end of the 90's \cite{feng_intelligent_1999} \cite{mahizhnan_smart_1999}. One of the first smart city was Singapore. Many others were following: Suwon, Seoul (Korea), Taipei (Taiwan), Mikita (Japan), Waterloo and Calgary (Canada), Glasgow (Scotland), New-York (USA) and Teheran (Iran), to name a few \cite{Talari_2017}. There are mainly three new software tools/technologies that allow smart cities to progress:  Big data, Internet of things (IoT) and artificial intelligence (AI). Talking about Big data and IoT, a study \cite{Cheng2015BuildingAB} present the case of Santander, Spain. A platform named "SmartSantander" has made this city one of the most connected in the world, with its 15,000 sensors (1,200 nodes) over its territory. Many sensors are statics and some others are mobile (mounted on some bus, taxis, or police cars, for instance). A big data platform named "CiDAP" was created especially for the city of Santander. Zampoglou et al. in \cite{Zampoglou_2014} present a good review of those useful technology for smart cities. Artificial intelligence is also a must for intelligent cities. Machine learning (ML) techniques like clustering \cite{Abed_2003} \cite{grekousis_modeling_2013} \cite{Foroutan_2012} \cite{arefiev_gis-based_2015}, neural networks \cite{Pijanowski_2009} \cite{Yang_2016}, Bayesian networks \cite{Kocabas_2013} \cite{liu_prediction_2014}, cellular automata \cite{CLARKE_1998} \cite{Rienow_2014} \cite{Yang_2016} and genetic algorithms \cite{Geroliminis_2011} \cite{Tong2016} \cite{Hao_2018} are very useful in this field. Learning from features of the past is the main strength of ML. Clustering allows to regroup similar features. Neural networks are useful to make some non-linear regressions. Bayesian networks can compute probabilities of some future event. Cellular automata allow to simulate a two dimensions geographical map. Finally, genetic algorithms (GA) are useful to optimize different types of configuration. 

\noindent In the large spectrum of smart cities is the intelligent urbanism \cite{benninger_principles_2002} \cite{santoso_intelligent_2013} \cite{murgante_geocomputation_2009} \cite{Wu_2010}. This specific part of smart cities aims to help urbanists to read better their city features and to predict how the urban territory will change. There are some paper studying different city district indexes. They all study some geographical region and geolocalized features from the past to predict what will happen in the future. The case of Attica (Athens area) in which they predict urban growth was proposed in \cite{grekousis_modeling_2013}. Their work aims to presents an artificial intelligence approach integrated with GIS for modeling urban evolution. They use a fuzzy logic system using a c-means clustering algorithm to divide the territory using fuzzy frontiers. In this system, each geographical position has a level of membership defined by a membership function \cite{Wang_Mendel_1992}. The clusters represent a specific level of urban growth. This system also uses a multilayers neural network (MNN) to learn and predict urban growth in the Attica area, by analyzing population changes over time and by building patterns. All the geographical data are managed by a GIS. Amongst many features, 9 has been selected to feed the system:  population, population growth in the decade, number of buildings, number of building growth in the last decade, use of residential sector, use of commercial sector, use of industrial sector, use of public sector, other uses. The results shown a clear profile for each district. For instance, for group ``A'', there is the strongest population growth rate (54.6\%). Buildings growth rate is also high at 96.5\%. The residential use of the land is high at 91.2\%.  We can conclude that group ``A'' represents a residential district in full growth. Data from the past was learned and used to predict growth rate of each district in the future. A very strong correlation between real and predicted features was shown.   

\noindent Some other papers are doing a similar work, but they model the territory using a cellular automaton \cite{Guan_Qingfeng_2005} \cite{Tayyebi_2011}. This is a good approach when features can be precisely geolocalized. In this case, it is possible to extract information and place it into a two-dimensional grid. From this grid, some simulation can be run according to some previously defined (or learned) rules. The results of those simulation can give some hints of what the territory will look like in the future. Obviously, the cellular automaton is used combined with some other artificial intelligence techniques to give some more complete results \cite{Li_Gar_2002} \cite{soltani_2013}.  

\noindent There is no standard for the evaluation and the prediction of a vitality index (VI). Some papers \cite{noauthor_there_nodate} \cite{noauthor_IDA_nodate} \cite{drewes_determining_2010} refer to VI, defining their own set of features (both qualitative and qualitative) and methods. Since each city does not archive the same data over the years, it is difficult to establish a standard set of features for the evaluation of a VI. Each city must use the consistent data available from the last decades. Being inspired by the previous works, this paper proposes a method based on ML algorithms to define, evaluate, and predict a VI in Trois-Rivi\`{e}res city. It shows the methodology for data preprocessing such as normalise and represent features, and to fill some gap. It defines how both supervised learning and unsupervised learning were used to calculate the VI and to make some prediction through years. The usage of a \textit{k}-means algorithm (unsupervised learning) that partially defines the index will be proposed. The usage of a feature-weighted inputs with stochastic gradient descent technique, will also be explained. Finally, we will see how a GA is used to optimize the clustering parameters. At the end, this proposed method based on ML algorithms provides some good insights for urbanists. \textit{}

\noindent The next sections of this paper are organized with the following structure: Section~2 describes the proposed methodology. Section~3 presents the results. Section~4 discusses about the results and their meaning and Section~5 concludes this research.

\section{The proposed method for the vitality index}%
\label{sec:Methodology}%
\subsection{Selected features}%
\label{subsec:Sub1_Selected_features}%
\noindent According to the Trois-Rivi\`{e}res urbanists, the ``vitality'' of a district can be defined by the strength of its economy, the health and social status of its citizen. Unlike the urban growth index, which tells if a territory is occupied by urban space, the VI index refers to the economic health of an urban territory and to the social condition of its citizens. This paper aims to define the VI, to evaluate it according to the Trois-Rivi\`{e}res city features and predict this index for each targeted district in the future. In urbanism context, we have access to massive data information. The first step was to prioritize and select the right features needed to calculate a VI. This was done in collaboration with urbanism experts. They have selected the features they thought could have a significant influence on a VI. Table  \ref{table_features} presents the eight selected features and the pre-processing applied on them.

\noindent ~~As shown in Table \ref{table_features}, pre-processing as been applied to each feature. First, every feature has been normalized using a MinMax function based on the assumption that each feature has same importance in the VI. Eq. \ref{eq:MinMax} shows the MinMax normalization formula.  It simply normalizes a number to get a 0 to 1 range, associating the smallest value to 0 and the highest to 1.  

\captionof{table}{Vitality index features and their pre-processing \label{table_features}}%
\noindent \includegraphics[width=\columnwidth]{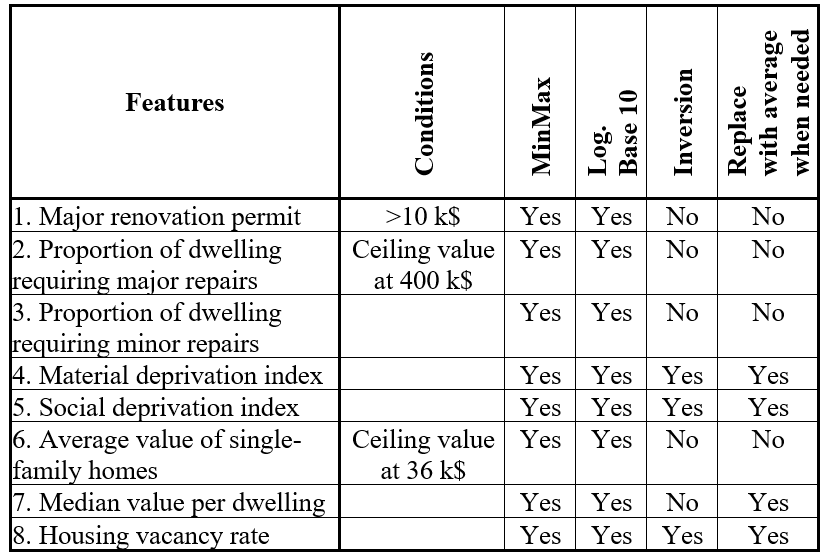}%

\noindent This is easier way to calculate the index and to present features on the same scale using different graphics. The presentation of the features is especially important since it must be interpreted by urbanists. To have a better feature distribution, a logarithmic function is used to scale the feature in logarithmic scale. Some features have been inverted to keep the consistency: 0 is always the worst feature value and 1 is always the better feature value. Finally, some average values were used when no data were available.

\begin{equation}%
z = \frac{x - min(x)}{max(x) - min(x)}%
 \label{eq:MinMax}%
\end{equation}%

\subsection{Framework design to predict vitality index}%
\label{subsec:Sub_Framework}%

\noindent The proposed framework design includes several parts to finally predict a VI. First, the model must learn from district's features the VI. Since the outputs of the past are unknown, an unsupervised learning technique (\textit{k}-mean algorithm) had to be used. This algorithm's parameters are optimized with a GA. Afterward, having all the VI for three years 2006, 2011 and 2016 in a 10 years range, the method based on this model architecture can predicted the district evolution in the future. This prediction is made using a linear regression. 

Fig. \ref{fig:architecture} shows the block diagram of the dataflow and the ML used represented in 3 steps: (1) GA and clustering, (2) for neural network, and (3) linear regression.\newline

\noindent \includegraphics[width=\columnwidth]{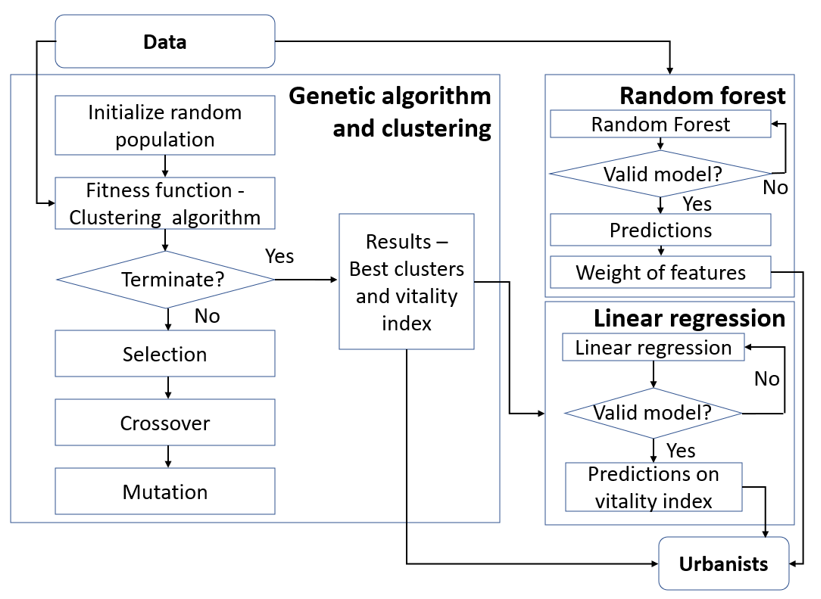}%
\captionof{figure}{Proposed framework design in three steps (1) genetic algorithm and clustering, (2) for neural network, and (3) linear regression.\newline \label{fig:architecture}}%

To have a whole system to evaluate and predict a VI, several types of ML algorithms must be used.

\subsection{Unsupervised learning –- k-means clustering}%
\label{subsec:Sub_kmeans}%

\noindent To determine the VI, it is necessary to use an unsupervised learning technique since there is no tagged label for each input data.  This algorithm will assign to each district a cluster reference letter, according to a similarity level of their features. A \textit{k}-means algorithm has been used to determine the clusters. 

\noindent Eq. \ref{kmeans} defined the \textit{k}-means clustering equation where \textit{J} is a clustering function, \textit{k} is the number of clusters, \textit{n} is the number of features, $x_{i}^{(j)}$ is the input (feature \textit{i} in cluster \textit{j}) and $c_{j}$ is the centroid for cluster \textit{j}. Centroids are obtained by randomly trying some values and selecting the best. 

\begin{equation} \label{kmeans} 
J=\sum _{j=1}^{k}\sum _{i=1}^{n}\left\| x_{i}^{(j)} -c_{j} \right\| ^{2}    
\end{equation} 

There are several metrics that allow to measure a clustering performance. Although, every metric is not compatible with every algorithm. Since a genetic algorithm was used to optimize the clustering (using several techniques), we had to make some choice according to the chosen clustering technique.  Since k-means algorithm was selected, the clustering performance has been measured by the ``Silhouette'' metrics. This metric is documented by Kaufman and Rousseeuw \cite{Rousseeuw_2009} and \cite{gueorguieva_mmfcm_2017}. This metric includes two important equations. The distance between each point and the center of its cluster is shown in Eq. \ref{kmeans}. The distance between the center of each cluster is shown in Eq. \ref{Silhouette1}. Finally, Eq. \ref{Silhouette3} uses the result of Eq. \ref{Silhouette1} and Eq. \ref{Silhouette2} to calculate the final "Silhouette score" that indicate the quality (the consistency) of the clustering. The silhouette ranges from -1 to +1. Values from -1 to 0 indicates that the point is associated to a wrong cluster and from 0 to 1 are associated to a good cluster. The higher the value, the better the cluster consistency \cite{Rousseeuw_2009}. 
\begin{equation} \label{Silhouette1} 
a(i)=\frac{1}{\left|c_{i} -1\right|} \sum _{j\in c_{i} i\ne j}d(i,j)  
\end{equation} 
\begin{equation} \label{Silhouette2} 
b(i)={\mathop{\min }\limits_{k\ne i}} \frac{1}{\left|c_{k} \right|} \sum _{j\in c_{k} }d(i,j)  
\end{equation} 
\begin{equation} \label{Silhouette3} 
s(i)=\frac{b(i)-a(i)}{\max \left(a(i),b(i)\right)} ,\; \; {\rm if}\; \left|C_{i} \right|>1 
\end{equation} 
If most elements have a high value, then the clustering configuration is appropriate. If many points have a low or negative value, then the clustering configuration may have too many or too few clusters. In our case, we had to create clusters of 5, 6, 7 or 8 dimensions. It is way more complicated to get a high silhouette score than with some 2- or 3-dimensions features. 

\subsection{Genetic algoriths}%
\label{subsec:Sub_Genetics}%

\noindent There are some relevant features to calculate a VI. Although, no label can be assigned to each set of features. Therefore, we can not use a supervised learning algorithm. Unsupervised learning algorithms allow a machine to learn without labels, though. There are several techniques to do so, each one using some different parameters. Consequently, there is a numerous of possible configurations. To optimize the results of the clustering, a GA (Cedeno, 1995) is used. Four genes are used in the evolution process: \textit{k}-mean maximum iteration parameter,\textit{ k}-mean \textit{n} centroid parameter, the number of clusters to find, and a list of features. Table \ref{table_genetique} shows the configuration of the GA. The fitness function was the silhouette score of the clustering. This metric that evaluate the cluster consistency is defined by Eq. \ref{Silhouette2}. The GA parameters are the following: 

\begin{enumerate}
\item  Number of generations: number of iterations on the fitness/breeding/mutation process.
\item  Chromosomes: Number of individuals configurations tested by the process. Population. 
\item  Initial chromosomes initialisation: The method used to initialize the chromosomes at generation 0.
\item  Mutation rate: At breeding time, a percentage of the chromosomes that do not inherits from parents but are randomly reinitialized.
\item  Percentage of chromosomes fitting well enough to breed: A threshold of the fitness function. The chromosomes ranking better than this value will be breeded in the next generation.
\end{enumerate}

\noindent The configuration maximizing the silhouette score is displayed in Table \ref{table_genetique} in the column "Best-Score", where we maximized the Silhouette score using GA considering different number of clusters, number of features among the 8 features (Table \ref{table_features}). We reach a Silhouette score of 0.46, with N\_${}_{init}$~=~14, Max\_${}_{iter}$~=~196, 5 clusters using the features 1, 6, 7, and 8. Otherwise, in some application cases, the number of clusters is fixed by the urbanism experts considering all features. 

\subsection{Feature-weighted inputs}%
\label{subsec:Sub_feature-weigthed_inputs}%

\noindent One important answer we had to find was the importance of each feature in the clustering process. To do so, a loop evaluating the totality of the feature's list combination has been processed. This clustering process returned a silhouette score for each feature combination. Having a list of features configuration and silhouette index, the "random forest regressor" technique was used to determine the importance of each feature. A random forest is an iteration over n "decision trees" (n = 250, in this case). The result was a list of importance ordered features and their weight. 

\subsection{Linear regression to predict VI}%
\label{subsec:Sub_lineair_regression}%

\noindent The last stage of the proposed framework concerns the prediction of VI in the future for each first surrounding area districts. The VI were available by dissemination area. There can be many dissemination areas in each district. We had to regroup them by district and calculate the linear regression line through the available years to predict 10 years later. Although, in some case, there is not many points in the cloud, and it is hazardous to conclude to a reliable prediction.

\section{Results applied on the first belt districts in
Trois-Rivières city}%
\label{sec:Results}%

\noindent The Trois-Rivi\`{e}res territory as we know it exists since an important fusion between six cities and municipalities, in 2002. In this case, data from before this fusion era is considered irrelevant. Trois-Rivi\`{e}res aera is 334 km${}^{2}$ and has 136 000 people living on its territory. 40\% of its territory is in agricultural area, 20\% in rural area and 40\% in urban area. It is situated at the junction of St-Laurent river and St-Maurice river, about mid distance from Montreal and Quebec City. Trois-Rivi\`{e}res is also known for its major infrastructures for planes, trains and ships. 

Fig. \ref{fig:citymap1} shows a map of Trois-Rivi\`{e}res. The greyed part is the first belt districts (the important part for this study). 

\noindent \includegraphics[width=\columnwidth]{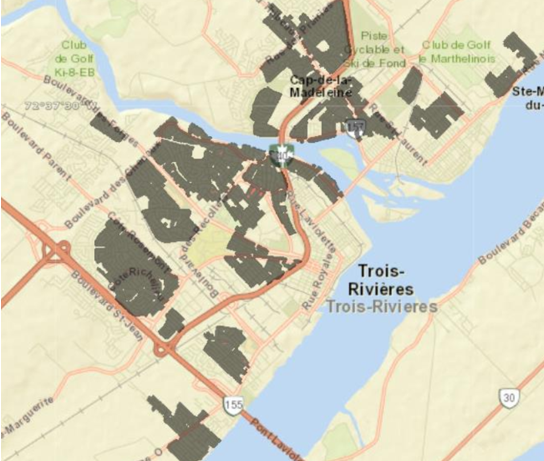}%
\captionof{figure}{The city of Trois-Rivières, Quebec, Canada. The grey
color defines the first belt area.\newline \label{fig:citymap1}}%

\noindent The urbanisation of its area happened in three steps. The first one was prior to 1950. This area is called ``firsts districts'' or ``central districts''.  The ``first belt'' or ``first agglomeration'' was built between 1950 and 1980.  Since then, the new areas are known to be the ``second belt'' area. Since the important fusion of 2002, residential development is more important than foreseen. 

\noindent The city of Trois-Rivi\`{e}res needed to have some insights to write its urbanism plan. Specially, there was a need to better foresee the vitality of the first belt districts.  The reason is that is some demographic issues (weak growth, aging of the population, etc.). The city needed to have more information about short-term vitality (5-10 years), average term vitality (10-20 years), and long-term vitality (20-30 years) of the first belt districts of Trois-Rivi\`{e}res. Basically, this study is focusing on this vitality aspect. Many more aspects may be studied in some future work. 

\subsection{Features distribution and representation}%
\label{subsec:Sub_Feature_rep}%

First, let us see the distribution of each of the 8 features for
year 2016. There is a similar distribution for each available
year. Section \ref{subsec:Sub1_Selected_features} (Table \ref{table_features}) show to methodology to obtain
these values. Fig. \ref{fig:feature_all} shows the distribution of the features. The X-axis represent the dissemination area (DA) which is a
geographical location in the city. There is 135 of them in
Trois-Rivières city. The Y-axis is the feature normalized
value.
For each of the 135 dissemination areas, and for every year,
all the features are represented on the same “radar” graphics.
Since all features are normalized, they can be displayed on
the same scale. Fig. \ref{fig:radar1} show an example of this radar graphics
(DA: 24370200 in year 2016).

\subsection{ Vitality index}%
\label{subsec: Sub_Vitality_index}%

In the graphics of Fig. \ref{fig:radar1}, we can extract the average of the
sum of the features. The result is also a normalized value
where a lower value means less vitality and a higher value
means more vitality. 

\noindent \includegraphics[width=\columnwidth]{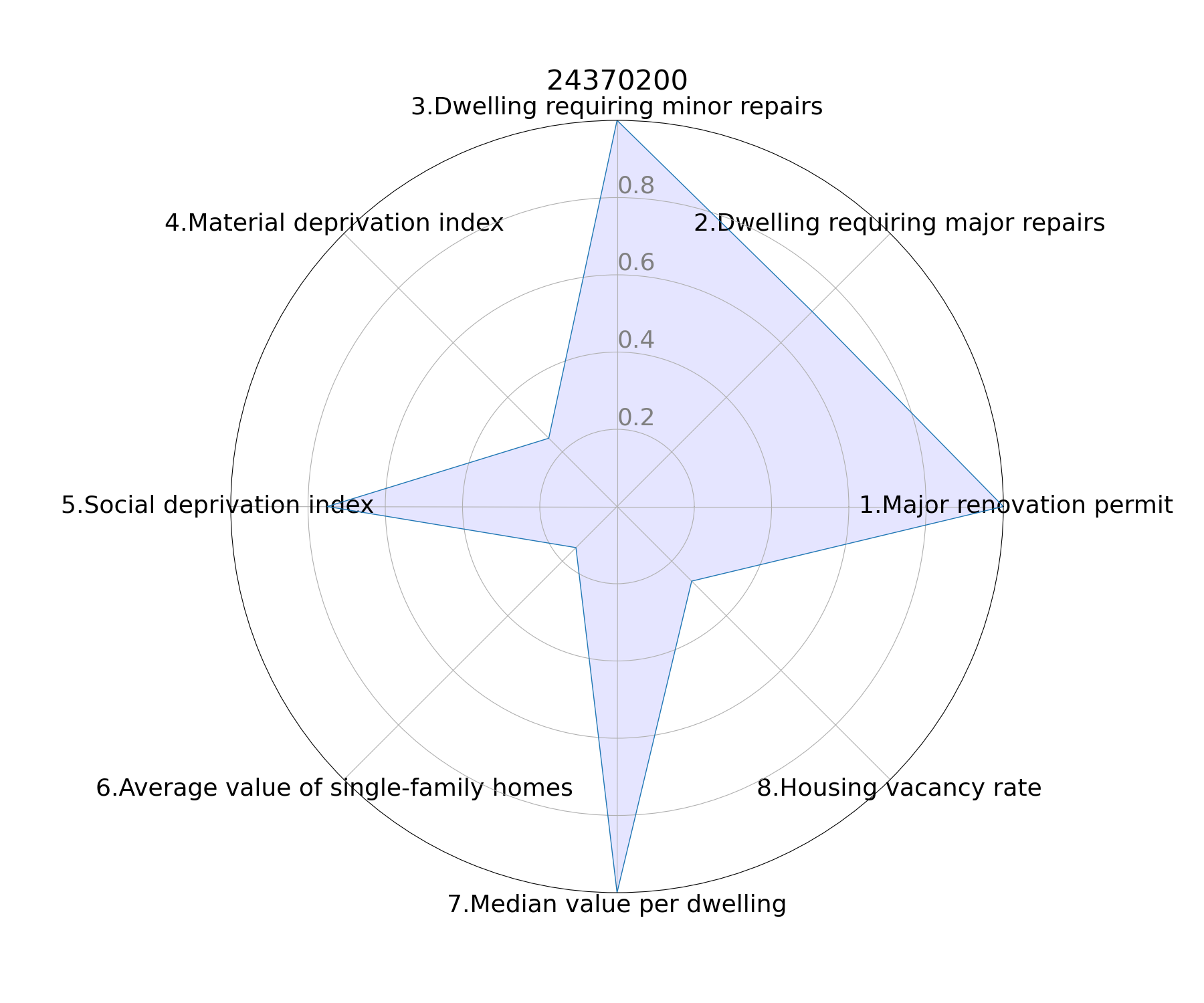}%
\captionof{figure}{Typical "Radar" graphic used to represent the 8-dimensional features.\newline \label{fig:radar1}}


\newpage
Feature 1 \newline
\noindent \includegraphics[width=6.50cm]{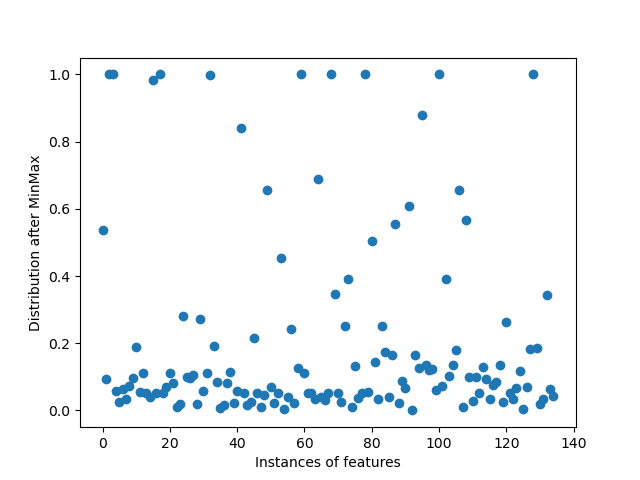}%

Feature 3 \newline  
\noindent \includegraphics[width=6.50cm]{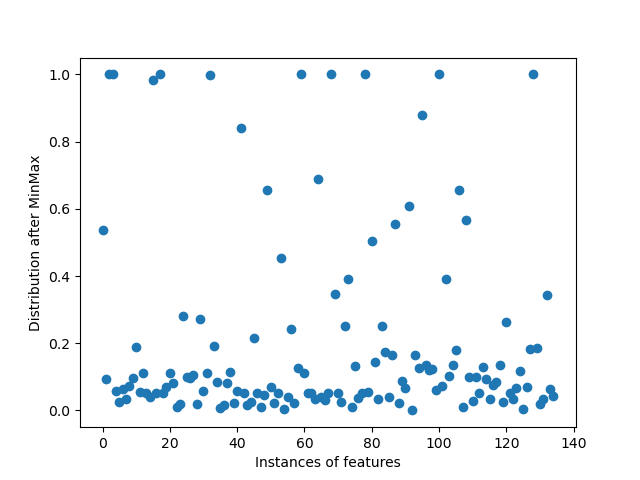}%

Feature 5 \newline  
\noindent \includegraphics[width=6.50cm]{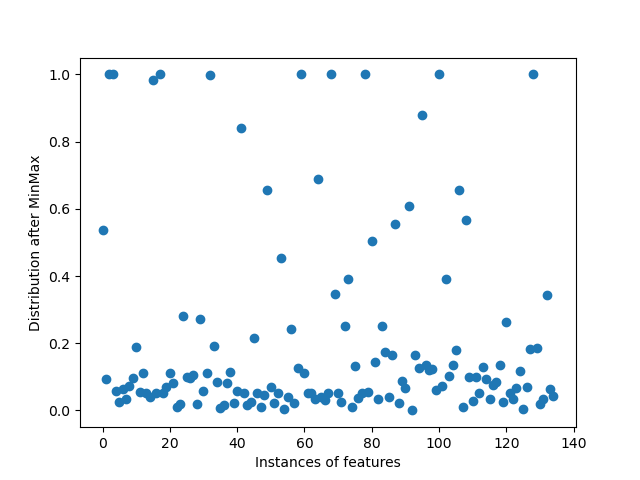}%

Feature 7 \newline  
\noindent \includegraphics[width=6.50cm]{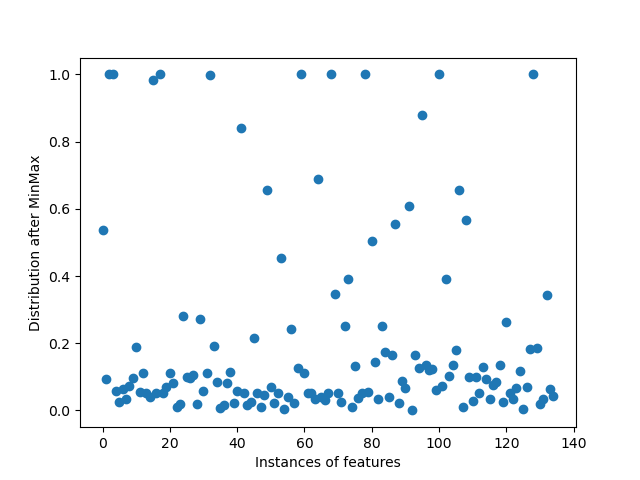}%

\vfill\null
\columnbreak

Feature 2 \newline  
\noindent \includegraphics[width=6.50cm]{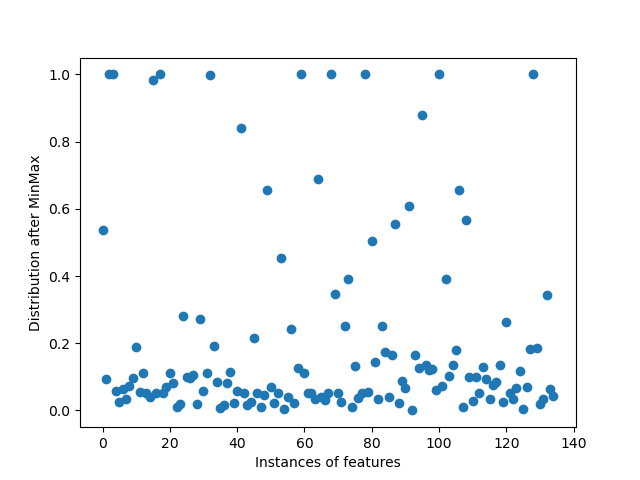}%

Feature 4 \newline 
\noindent \includegraphics[width=6.50cm]{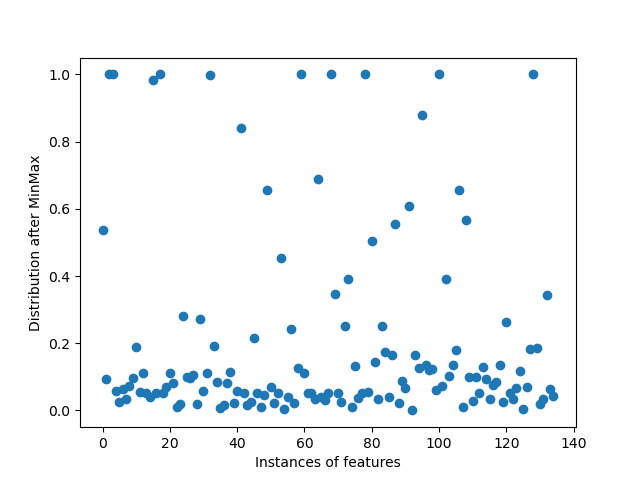}%

Feature 6 \newline  
\noindent \includegraphics[width=6.50cm]{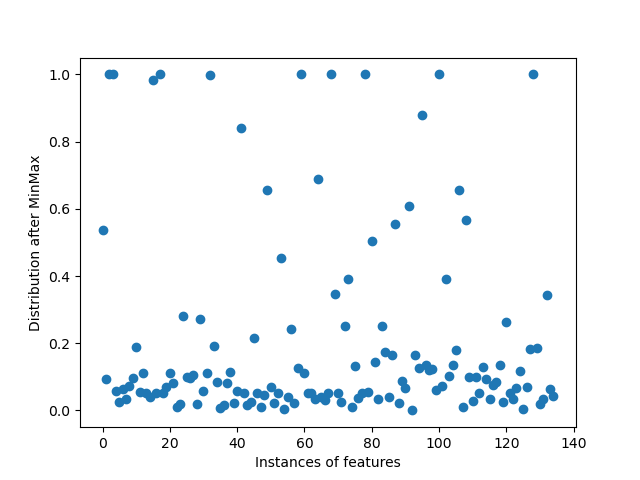}%

Feature 8 \newline  
\noindent \includegraphics[width=6.50cm]{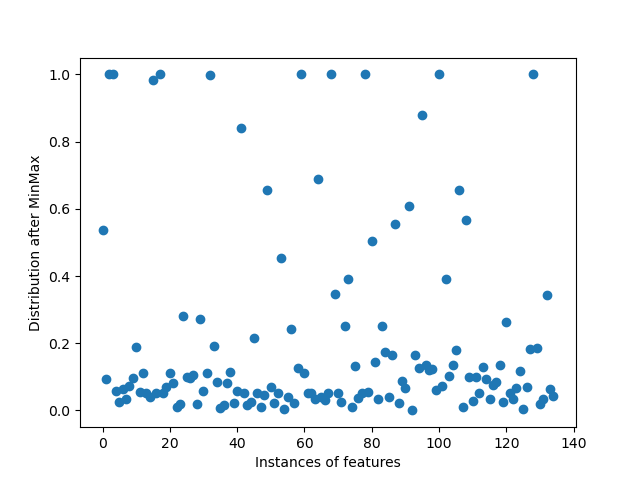}%

\noindent \includegraphics[width=0cm]{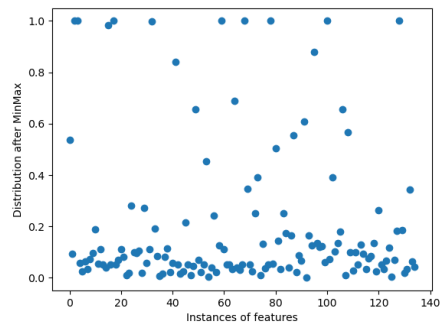}%
\noindent \captionof{figure}{Features 1 to 8 \label{fig:feature_all}}%
\newpage

Although, this information is incomplete. There are some very different district profiles having the same average of the sum of the features. The best
way to visualise the district profile is to regroup them. This
was made by using the clustering technique described in
section \ref{subsec:Sub_Clustering} For this reason, this research defines the vitality index by the two parts, as following:

\begin{enumerate}
\item  A letter representing the profile (cluster), and
\item  A number representing the average of the sum of the
features.
\end{enumerate}

For instance, "C45" means a "C" cluster with an average of
the sum of the features of 0.45. The 45 value is the average
sum of the feature (0.45) multiplied by 100. The inspiration
of this classification system comes from the works on star
classification by Annie-Jump Cannon in \cite{Cannon_1912}. In this two-dimension notation system, a star
could have a G5 type.

\subsection{Clustering}%
\label{subsec:Sub_Clustering}%

Like described in Section \ref{subsec:Sub_Clustering}, the number of clusters and the
number of used features has been determined by thousands of
simulations of GA including feedback analysis from
urbanists. We were to use all the 8 features and to divide the
135 aeras of dissemination in 10 clusters. The distribution of
the clusters (year 2016) in shown in Fig. \ref{fig:kmean1}. For each cluster
on the X-axis, a distribution level on the Y-axis.
Fig. \ref{fig:kmean2} shows on the Y-axis the average of the sum of each
feature (year 2016), for each dissemination area (X-axis).
Vertical red lines divide the clusters, and horizontal dotted
lines show the average of each cluster.
The best way to visualize and interpret every cluster is to
superpose every radar graphic of dissemination area of the
same group. In Fig. \ref{fig:kmean2}, we can see that the 6th and 7th
(clusters F and G) have about the same average of the sum of
the features (around 0.4). Without having their cluster profile,
it would be impossible to see the difference. 

Fig. \ref{fig:radarmulti1} and Fig. \ref{fig:radarmulti2} show the profile of those two clusters. We can easily see that even though they both have a similar average of the sum of the features, the distribution of theses sum values is not the same. They have a very different profile.

\noindent \includegraphics[width=\columnwidth]{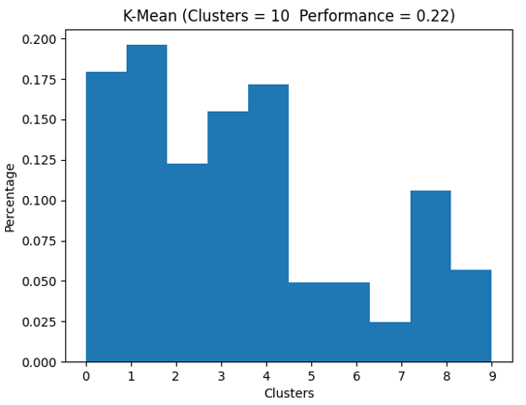}%
\captionof{figure}{Clusters distribution histogram\newline \label{fig:kmean1}}%

\noindent \includegraphics[width=\columnwidth]{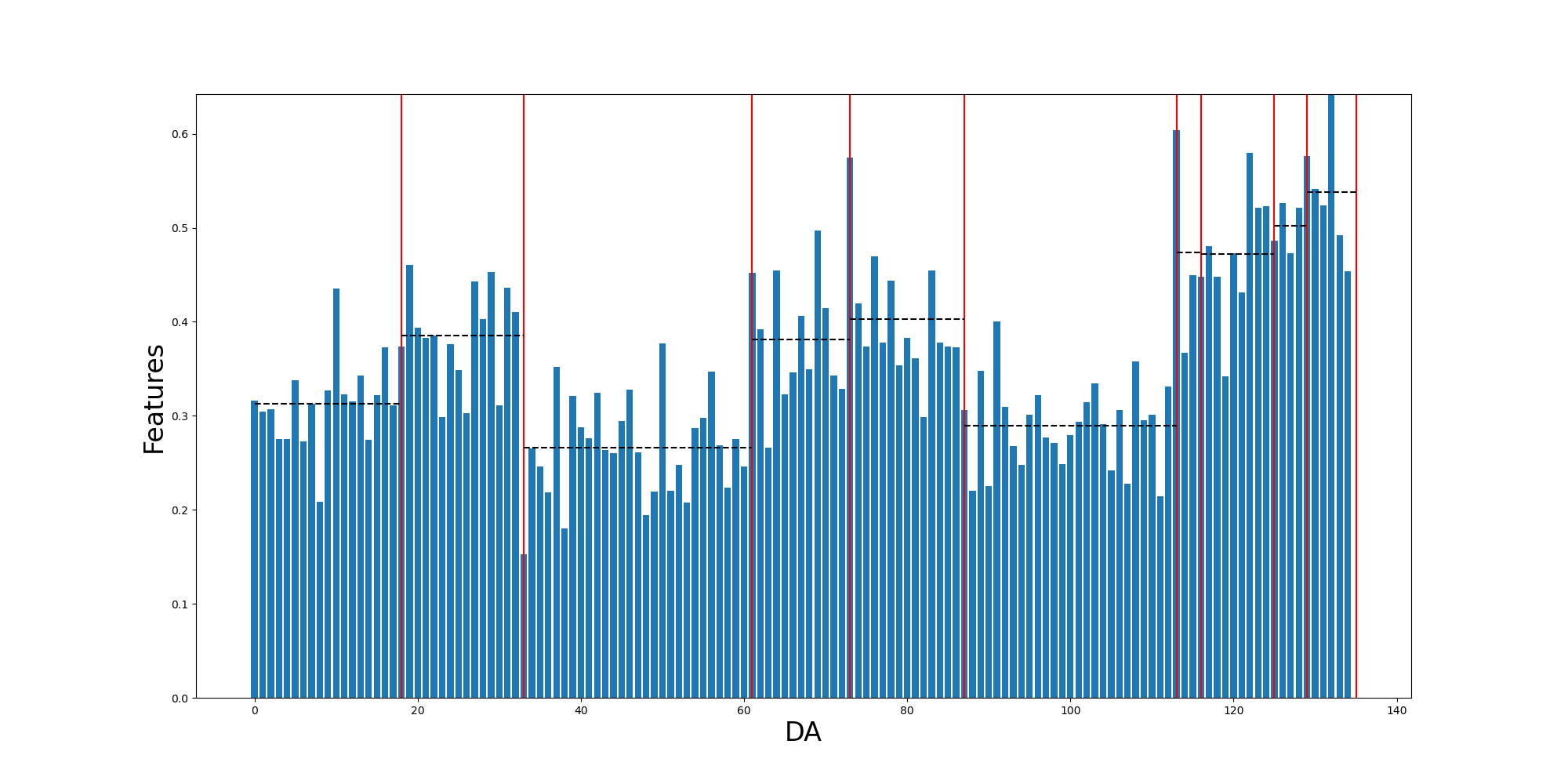}%
\captionof{figure}{Feature average for each dissemination area, cluster
division and cluster average\newline \label{fig:kmean2}}%

\noindent \includegraphics[width=\columnwidth]{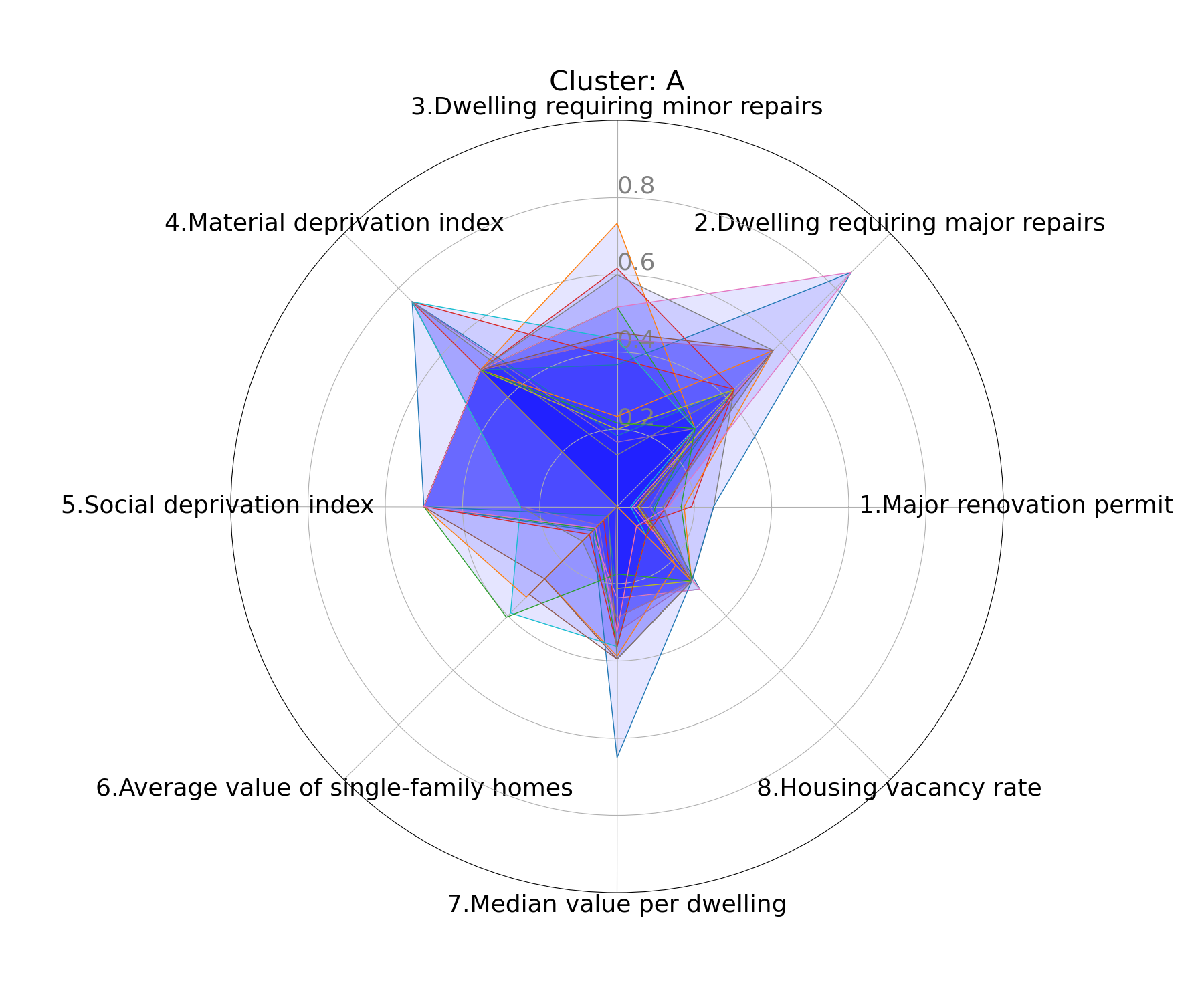}%
\captionof{figure}{Cluster A and its stacked radar graphics\newline \label{fig:radarmulti1}}%

It is even easier to see when both the profile and average of
the sum of the features are different. Fig. \ref{fig:radarmulti3} shows cluster I.
Most of the cluster members are rather smalls. 

\noindent \includegraphics[width=\columnwidth]{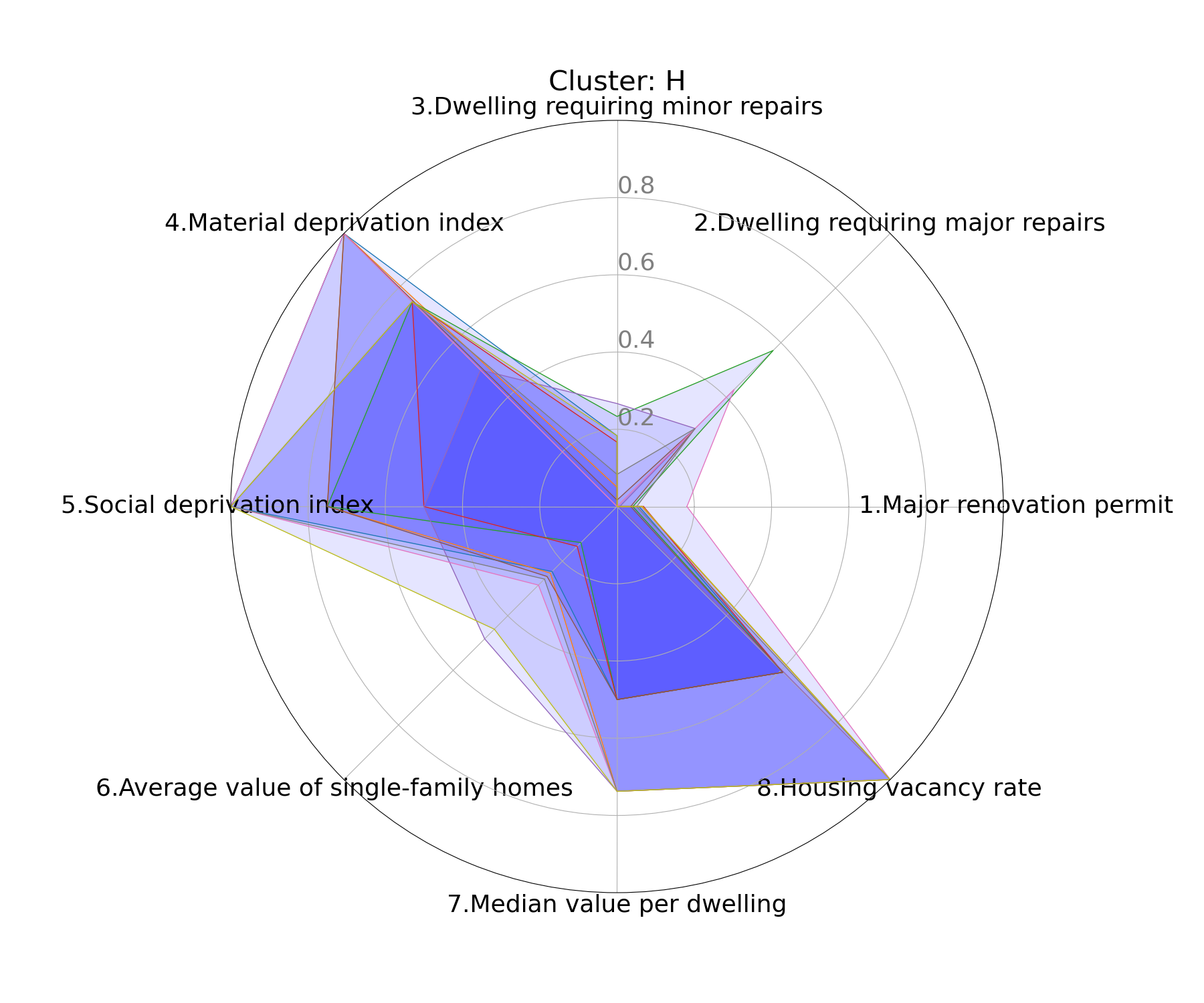}%
\captionof{figure}{Cluster H and its stacked radar graphics\newline \label{fig:radarmulti2}}%

\noindent \includegraphics[width=\columnwidth]{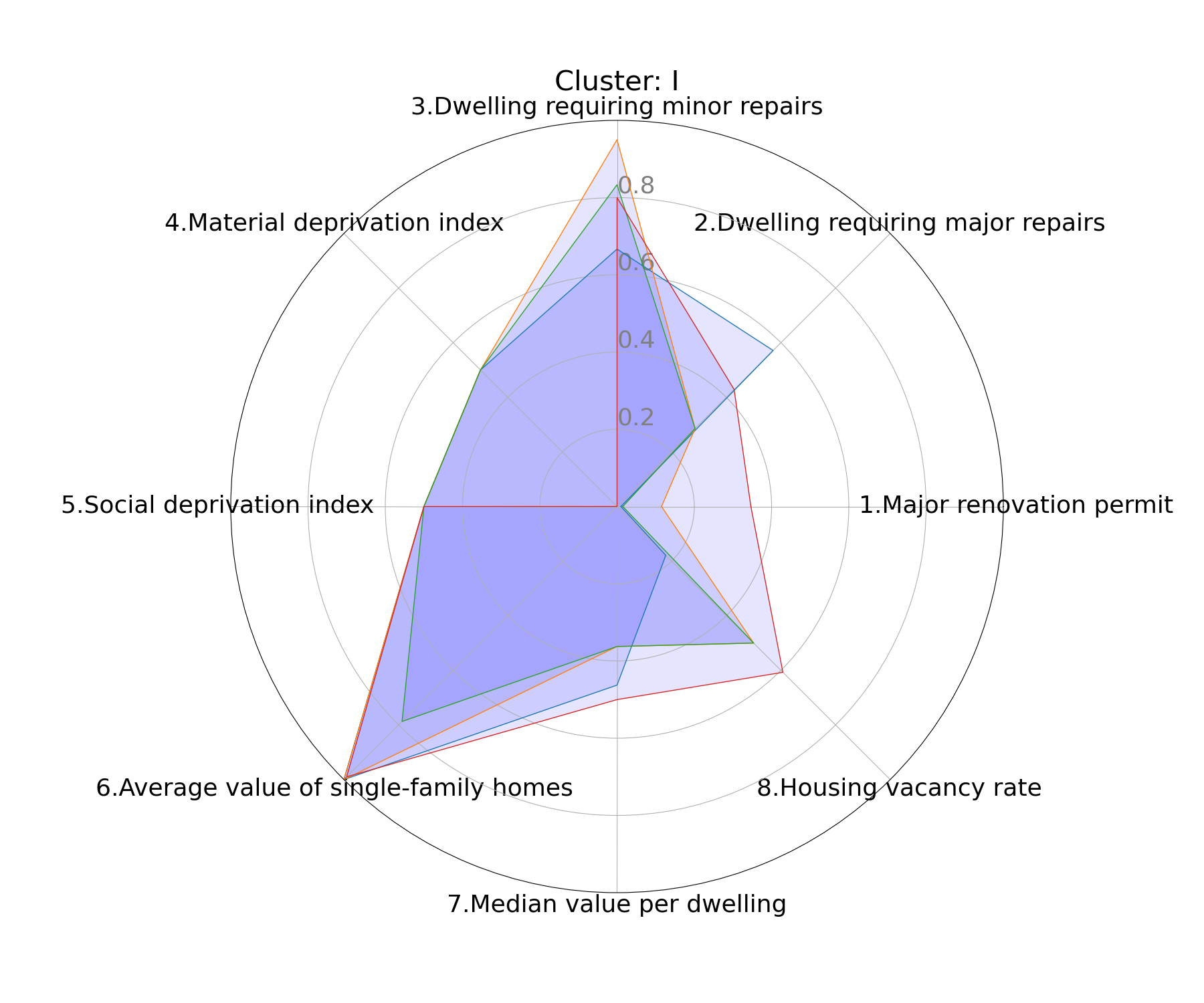}%
\captionof{figure}{Cluster I and its stacked radar graphics\newline \label{fig:radarmulti3}}%

Fig. \ref{Silhouette1}a shows the distribution of the features between 5
clusters. We can note that there are very few clustering errors
(from -1 to 0) and a silhouette score value of 0.46. This is the
configuration that optimizes the silhouette score.
Fig. \ref{Silhouette1}b shows the distribution of the features between 10 clusters.  We can note that there are very few clustering errors (from -1 to 0) and a silhouette score value of 0.22. 

As mentioned earlier, the clustering process was optimized by a GA. The results shown in Table \ref{table_genetique} for 10 clusters given Silhouette score of 0.22. The N${}_{\_}$${}_{init}$ parameter is the number of times the k-means algorithm will be run with different centroid seeds. The Max${}_{\ }$${}_{iter}$ is the maximum number of iterations of the k-means algorithm for a single run.

\noindent \includegraphics[width=\columnwidth]{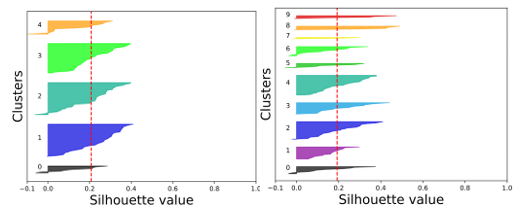}%
\captionof{figure}{Silhouette metrics using 4 features: a) 5 clusters
b) 10 clusters (2016 data).\newline \label{fig:Silhouette1}}%

\noindent Fig. \ref{fig:genetique} shows the typical evolution of fitness function of GA defined by the Silhouette score. Red curve represents the silhouette score average and green curve represents the silhouette score maximum.

\captionof{table}{Chromosome clustering configuration maximizing the Silhouette score
(Best-Cluster) and specifying the number of cluster (Fixed-Cluster).\newline \label{table_genetique}}%
\noindent \includegraphics[width=\columnwidth]{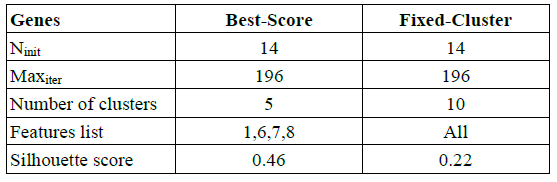}%

\noindent \includegraphics[width=\columnwidth]{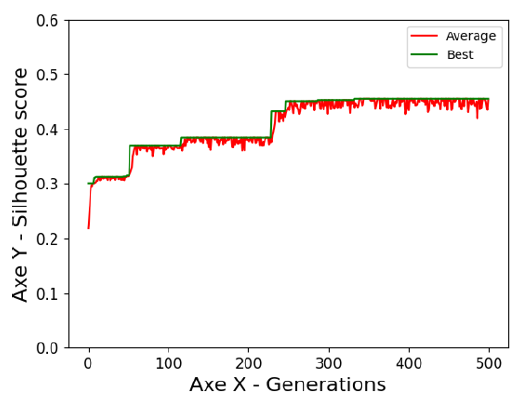}%
\captionof{figure}{Typical evolution of silhouette score over generations.\newline \label{fig:genetique}}%

\subsection{Weighting the features}%
\label{subsec:Sub_Weigthing_features}%

Urbanists wanted to know which features are the most
relevant in the clustering process. Section \ref{subsec:Sub_feature-weigthed_inputs} presented a
methodology based on a Random Forest algorithm to weight
the 8 features proposed by urbanists. The weights of the
features in 2016 are given by Fig. \ref{fig:RF}.
In order of importance, from the most important to the least
important (the number in parenthesis is the level of
importance): 

\begin{enumerate}
\item feature 3 (0.125) Proportion of dwelling requiring minor
repairs
\item feature 2 (0.122363) Proportion of dwelling requiring major
repairs
\item feature 8 (0.113767) Housing vacancy rate
\item feature 6 (0.112084) Average value of single-family
homes
\item feature 5 (0.109790) Social deprivation index
\item feature 4 (0.109603) Material deprivation index
\item feature 7 (0.109569) Median value per dwelling 
\item feature 1 (0.109211) Major renovation permit
\end{enumerate}

\noindent \includegraphics[width=\columnwidth]{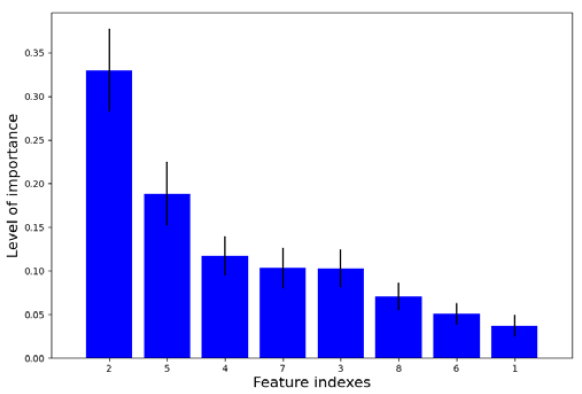}%
\captionof{figure}{Weights of feature.\newline \label{fig:RF}}%

\subsection{Predicting vitality indexes}%
\label{subsec:Sub_pred_VI}%

This research is only able to predict the average of the sum of
the features part of the VI.
At least for the numeric part it is possible to have a regression
that learns from the past to estimate future. Since the goal is
to predict vitality in the first surrounding area, we have first
to regroup dissemination area. Fig. \ref{fig:citymap2} shows a district that includes four areas of dissemination. It is usually from 1 to 7
per district.
To make some prediction, we must plot the VI (numeric part)
of each aera of dissemination included in each district. Then
the regression line must be added and used to make the
prediction about the future. In this case, results must be
interpreted with caution since there are only three years of
history to predict years 2021 and 2026. Fig. \ref{fig:regression} shows an
example of such a prediction using past VI (numeric part)
computed on 6 areas of dissemination (on 3 years 2006, 2011
and 2016). Each aera of dissemination is defined by a blue
dot. In this case, the 6 areas are those included in the TR-3
district.
Those results are some examples taken in the final report
\cite{Dessureault_2019} written for
urbanists of the city of Trois-Rivières. Obviously, this report
includes hundreds of figures to illustrate data on every district
of the first belt and every dissemination area. 

\noindent \includegraphics[width=\columnwidth]{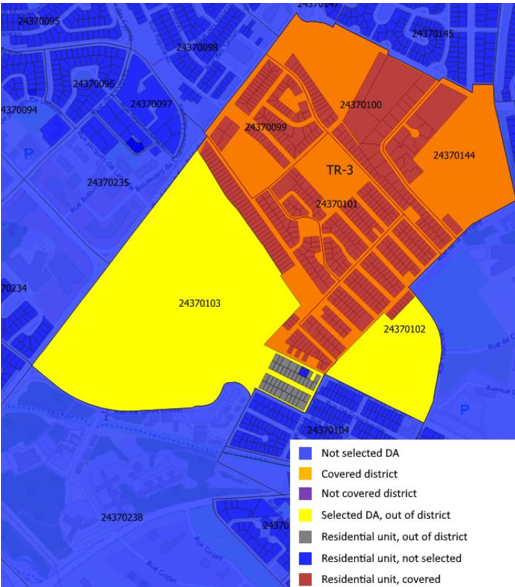}%
\captionof{figure}{Map of TR-3 district of Trois-Rivières and its
dissemination areas.\newline \label{fig:citymap2}}%

\noindent \includegraphics[width=\columnwidth]{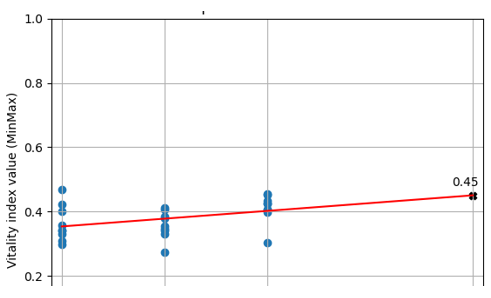}%
\captionof{figure}{Regression line and vitality prediction based on
areas of dissemination of the past, for TR-6 district of TroisRivières city.\newline \label{fig:regression}}%

\section{Discussions}%
\label{sec:Discussions}%
A method for calculating and predicting a VI has been
developed in this paper. The main objective was to help
urbanists to have a better understanding of the raw data
available in different sources, including their own. The data
was collected, then pre-processed to optimize distribution and
readability. Four ML algorithms were used to process data:
k-mean clustering technique, GA, feature-weighted inputs
and linear regression. There is no simple way to use a GA on
some clustering techniques. For different clustering
techniques, we must use different types of parameters. There
are also some issues about the evaluation of the results of the
clustering. The Silhouette metrics finally did a fair job to
evaluate the clusters consistency. Obviously, the clustering of
some 8-dimensions indexes is a greater challenge than the 2-
dimensions points cloud clusters usually presented. Due to
this higher dimensionality, there was also some issues about
the graphical representation of the features. The “radar”
graphic type was very useful. The new custom “Omni”
graphic type invented for the purpose was also helpful to
present the totality of the information at a glimpse. The
superposed radar graphic was also helpful to visualize the
clusters consistency, maybe in an even better way than using
the Silhouette metric. The are some tables of appreciation of
the Silhouette scores, but they are based on some 2-
dimensions features. It is very difficult to know if some 8-
dimensions features clusters (like the ones in this research)
are consistent or not. That is why the superposition of the
radar graphics was so important to confirm the clusters
consistency.
At the end, this research succeeds in converting some
scattered raw data in some valuable knowledge, well
presented and useful for the writing of the urban plan of
Trois-Rivières city. 

\section{Conclusion}%
\label{sec:Conclusion}%
All the code written in this research has some great
generalization perspectives. In a near future, it could be
converted in a more general urban tool to make some
prediction about a broader range of urban indexes, such as
criminality indexes, health indexes or economic indexes.
There are also some improvement possibilities in the
clustering part. The clustering method and the metrics could
be studied and improve. One way of improving a next version
would be to replace the k-mean clustering algorithm by a
c-mean algorithm. This would have the benefits of fuzzifying
the districts limits. A model based on fuzzy clustering would
reflect reality in a better way than a model based on a crispy
clustering algorithm. Some improvement can also be done by
making some prediction on each available feature, instead of
only the numeric part of the VI (which is the average of those
features). Since this system must deal with an input that
includes multiple features, some algorithms based on
dimensionality reduction must be explored. There could be
some improvement possibilities by using this solution in the
pro-processing phase. Finally, in a next version, it will be
easy to also project the profile part of the VI. It will be
certainly possible to predict the shapes of the multidimension VI of the future.

\section{Acknowledment}%
\label{sec:Acknowledment}%
This work has been supported by the City of Trois{-}Rivières, The "Cellule d’expertise en robotique et intelligence artificielle" of the Cégep de Trois{-}Rivières and IDE Trois{-}Rivières.

\bibliographystyle{plain}%
\bibliography{paper}

\end{multicols}%
\end{document}